\documentclass[runningheads, a4paper]{llncs}

\usepackage{amssymb}
\usepackage{graphicx}
\usepackage{times}
\usepackage[applemac]{inputenc}
\usepackage[T1]{fontenc}
\usepackage{prooftree}

\input{qobitree}

\newcommand{\resp}{\textit{resp.}~}
\newcommand\seq\vdash 
\newcommand\lto{\mathbin{\backslash}}
\newcommand\lfrom{\mathbin{/}}

\title{Encoding Phases using Commutativity and Non-commutativity in a Logical Framework}
\titlerunning{Phases in Logical Framework}
\author{Maxime Amblard \inst{}
}

\institute{
LORIA - INRIA Nancy Grand Est - BP 239 - 54506 Vandoeuvre-lès-Nancy Cedex \\
 Université Nancy 2,13 rue Maréchal Ney - 54037 Nancy cedex \\
 INPL, 2 av. de la Forêt de Haye - BP 3 - F-54501 Vandoeuvre
\\ \email{amblard@loria.fr}
}

\begin{document}

\maketitle

\begin{abstract}This article presents an extension of Minimalist Categorial Grammars (MCG) to encode Chomsky's \textit{phases}. These grammars are based on Partially Commutative Logic (PCL) and encode properties of Minimalist Grammars (MG) of Stabler \cite{Sta97}. The first implementation of MCG were using both non-commutative properties (to respect the linear word order in an utterance) and commutative ones (to model features of different constituents). Here, we propose to augment Chomsky's \textit{phases} with the non-commutative tensor product of the logic. Then we can give account of the PIC \cite{NC99} just with logical properties of the framework instead of defining a specific rule.
\end{abstract}

Generative theory has undergone many changes since Chomsky's Syntactic Structures \cite{Cho57} leading up to what is the Minimalist Program (MP) \cite{Chom95}.
The most frequent criticism made is certainly the non-computational and non-formal nature of such an approach.
It is nevertheless rich of a vast literature for the linguistic approach.
Fundamentals to the MP are the description of a main calculus which takes into account the syntax, and the production of two forms: one supposed to reflect the sequence of words, and another one for the semantic structure of the utterance.
Following the MP, Chomsky claims the identification of \textit{phases} in the syntactic derivation \cite{NC99}.
The verb, the main driving force of the analysis, is being transformed, opening the possibility of specific modifications, especially the definition of the Phase Impenetrability Condition (PIC).

The first proposal of formalization for MP was made by Stabler in \cite{Sta97}.
However, this formulation is far from the usual Montagovian approach of semantics.
Therefore, translations of this formulation into logic systems have been proposed, in particular \cite{LR01acl}, \cite{AL05}.
Much has been done, exploiting Curry’s distinction between the tectogrammatical and the phenogrammatical levels, and this has led to interesting proposals \cite{GM94}, \cite{deG01}, \cite{RM03}, \cite{CP07}.
The latest proposals of extension defined the MCG based on a fragment of Partially Commutative Logic (PCL) \cite{AM07th}, \cite{ALR10}.
The authors highlighted the  simultaneous need of commutative and non-commutative properties to produce  a useful framework .

In this paper, we propose a reconsideration of the properties of commutativity and non-commuta\-ti\-vi\-ty in MCG to account  the concept of \textit{phase} introduced by Chomsky.
Due to space consideration, we will limit ourselves to the problems of parsing, leaving aside the semantic aspects.
However, it should be noted that the syntax-semantics interface of MCG contains all the necessary material for its integration.

We first discuss the need for the two different relations in MCG. Then, we define Minimalist Categorial Grammars (MCG). Based on these definitions, the third section presents and encodes \textit{phases} in MCG, and shows the implementation of the PIC using only logical properties.

\section{Commutativity vs Non-Commutativity in Standard MCG and Phases}

To link logic and Generative Theory, MCG's derivations are proofs of a restriction of Partially Commutative Logic (PCL),
\cite{Ret97tlca}, seen as syntactic representations in generative theory.
This logic is an extension of Lambek calculus containing  simultaneously commutative and non-commutative connectives (\textit{ie} introduction and elimination of implication and tensor). To handle the different relations between hypotheses in the same framework, an entropy rule (restriction of order) is added.
Moreover,\cite{MR07} shows a weak normalization of this calculus, to produce regular analysis in MCG.

All definitions of these grammars are given in \cite{AM07th}\footnote{These definitions contain also a syntax / semantic interface. However, we leave this part out of this article due to space.}, with a composition of rules (note that according to these definitions, normalisation is strong). 
Moreover, this restriction does not use introduction of hypothesis.
They appear in the derivation only from specific lexical entries: in the hypotheses of a given category and that category as a formula. The lexicon will contain the entry $ \phi \seq \phi $ with $ \phi $ as a given category.

The concrete part of the proof is achieved by a \textit{merge} whose heart is the elimination of $ \lfrom $ or  $ \lto$ (rules that are found in different versions from categorial grammars ,
\cite{JL58}, \cite{MS87} \cite{Moo97}) plus the  entropy rule. 
This is one point where commutativity and non commutativity play a crucial role. In particular, for the word order, it is clear that the relation is non-commutativite: being on the right or the left of a given word could not just be the same.
Non-commutativity is also needed here, because of the second rule in MCG.

The hypotheses are seen as special position markers in the sequence of hypotheses of the proof.
 They are considered as resources of prominent features related to a phrase.
 They are unloaded by using the \textit{move} operation of the generative theory.
A direct implication is that the sequence of hypotheses in the proof contains exactly the sequence of available resources for further derivation.
 In this case, this sequence is a collection of resources and could not be a strict list.
 Unless a canonical order on the sequence of applied rules is presupposed, the only way to express this is with non-commutativity.
The \textit{Merge} rule must contain a release of the order.
Definitions of basic rules of MCG involve commutativity and non-commutativity.
We will show how we could use these properties in another perspective to encode a linguistic concept, more precisely by noting that the changing category of the verb controls the process flow.

Chomsky's theory introduces the notion of \textit{phase}, which corresponds to the evolution of the verb.
 Thus, we assume that the lexical item associated with the verb carries only part of its achievement in the sentence. This is an important use in the syntax-semantic interface for reporting UTAH \footnote{Uniform Theta Assignment Hypothesis or assignment of thematic roles.}. 
The needed hypothesis to introduce a DP is provided by verb lexical items.
And this is the correspondence of verb's resources with the DP category which allows the DP to be in the proof.

In the \textit{phase}'s definition, Chomsky assumes that some instance of \textit{move} (unification of features in our formalism) must be completed before the end of the \textit{phase}. Once it is reached, the specific resources of the process are no longer accessible to the rest of the analysis. It defines an \textit{island}  in the analysis, called Phase Impenetrability Condition (PIC).
 Translating this definition into our formalism implies that a \textit{phase} (or its representation) block access to part of the sequence of hypotheses of the proof. We assume that the interpretation in MCG is a non-commutative point in the sequence of hypotheses. 
 
The direct implication is that the analysis of a sentence simultaneously uses relations to link noun and verb phrases, and non-commutative relations to construct the analysis (changing the category of the verb). The use of non-commutativity involves a strict order to control the verb's role in the analysis. 
Formal properties used by items of different categories are disjointed and it allows to fully exploit their relations.

\section{Minimalist Categorial Grammars}

Minimalist Categorial grammars are based on a fragment of PCL to encode the MP of Chomsky. Then it uses an abstract calculus to produce both a string (sequence of words) and a semantic representation (formula). Word order and semantics are synchronized over the main calculus which uses a structured lexicon and specific rules.
First, we briefly present rules of PCL, then we introduce labels which encode word order, then we define lexical items and finally introduce rules of MCG. 

\subsection{Partially Commutative Logic (PCL)}

The logic introduced in \cite{Gro96} and extended in \cite{retore04} is a superimposition of the Lambek calculus (Intuitionistic Non-Commutative Multiplicative Linear Logic) and Intuitionistic Commutative Multiplicative Linear Logic.
Connectives are:

\begin{itemize}
\item the Lambek calculus connectives: $\odot$, $\lto$ and $\lfrom$ \hfill (non-commutatives)
\item commutative multiplicative linear connectives: $\otimes$ and $\multimap$ \hfill (commutatives)
\end{itemize}

The use of these connectives is presented in figure \ref{reglesmixtes}.

In this logic, commutative and non-commutative relations could be used simultaneously.
Then contexts are partially ordered multisets of formulae and in order to relax this order, we need an \textit{entropy} rule noted $\sqsubset$ which is defined as the replacement of $;$ by $,$.
The restriction of elimination rules and the entropy rule is called Minimalist Logic (the logic used to define MCG). In the following, we note $F$ the set of categories ($F$ stands for features).

\begin{figure}
\begin{small}

\prooftree
	\Gamma \vdash A
	\quad \Delta \vdash  A \backslash C
	\justifies
	<\Gamma; \Delta> \vdash C 
	\using
             [\backslash_{e}]
         \thickness=0.07em
\endprooftree
\hfill
\prooftree
	\Delta \vdash  A / C
	\quad \Gamma \vdash A
	\justifies
	< \Delta ; \Gamma> \vdash C 
	\using
             [/_{e}]
         \thickness=0.07em
\endprooftree
\hfill
\prooftree
	 \Gamma \vdash A
	\quad \Delta \vdash  A \multimap C
	\justifies
	(\Gamma, \Delta) \vdash C 
	\using
             [\multimap_{e}]
         \thickness=0.07em
\endprooftree

\bigskip 

\prooftree
	<A; \Gamma> \vdash C
	\justifies
	\Gamma \vdash A \backslash C 
	\using
             [\backslash_{i}]
         \thickness=0.07em
\endprooftree
\hfill
\prooftree
	<\Gamma; A> \vdash  C
	\justifies
	\Gamma \vdash C/A 
	\using
             [/_{i}]
         \thickness=0.07em
\endprooftree
\hfill
\prooftree
	(A, \Gamma ) \vdash  C
	\justifies
	\Gamma \vdash A \multimap C 
	\using
             [\multimap_{i}]
         \thickness=0.07em
\endprooftree

\bigskip 

\prooftree
	\Delta \vdash  A \odot B
	\quad \Gamma, <A; B>, \Gamma ' \vdash C
	\justifies
	\Gamma, \Delta , \Gamma '  \vdash C 
	\using
             [\odot_{e}]
         \thickness=0.07em
\endprooftree
\hfill 
\prooftree
	\Delta \vdash  A \otimes B
	\quad \Gamma, (A, B),  \Gamma ' \vdash C
	\justifies
	\Gamma, \Delta , \Gamma '  \vdash C 
	\using
             [\otimes_{e}]
         \thickness=0.07em
\endprooftree

\bigskip 

\prooftree
	\Delta \vdash A
	\quad \Gamma \vdash B
	\justifies
	<\Delta; \Gamma> \vdash A \odot B
	\using
             [\odot_{i}]
         \thickness=0.07em
\endprooftree
\hfill
\prooftree
	\Delta \vdash A
	\quad \Gamma \vdash B
	\justifies
	(\Delta, \Gamma) \vdash A \otimes B
	\using
             [\otimes_{i}]
         \thickness=0.07em
\endprooftree

\prooftree
	\justifies
	A \vdash A
	\using
             [axiom]
         \thickness=0.07em
\endprooftree
\hfill 
\prooftree
	\Gamma \vdash C
	\justifies
	\Gamma' \vdash C 
	\using
             [\mbox{entropy --- whenever }\Gamma'\sqsubset\Gamma]
         \thickness=0.07em
\endprooftree
\end{small} 
\caption{Partially Commutative Logic rules\label{reglesmixtes}.}
\end{figure}

\subsection{Labels encoding word order}

Derivations of MCG are labelled proofs of the PCL.
Before defining labelling, we define labels and operations on them. To do this, we use the set of phonological form $Ph$ and a set $V$ of variables such that: $Ph \cap V = \emptyset $.
We note $T$ the union of $Ph$ and $V$. We define the set $\Sigma$, called \emph{labels set} as the set of triplets of elements of $T^\ast$. Every position in a triplet has a linguistic interpretation: they correspond to specifier/head/complement relations of minimalist trees.
A label $r$ will be considered as $r=(r_{s}, r_{h}, r_{c})$.

We introduce variables in the string triplets and a substitution operation. They are used to modify a position inside a triplet by a specific material. Intuitively, this is the counterpart in the phonological calculus of the product elimination.

A \emph{substitution} is a partial function from $V$ to $T^\ast$.
For $\sigma$ a substitution, $s$ a string of $T^{\ast}$ and $r$ a label, we note respectively $s.\sigma$ and $r.\sigma$ the string and the label obtained by the simultaneous substitution in $s$ and $r$ of the variables with the values associated by  $\sigma$ (variables for which $\sigma$ is not defined remain the same).

If the domain of definition of a substitution $\sigma$ is finite and equal to $x_1, \dots, x_n$ and $\sigma(x_i) = t_i$, then $\sigma$ is denoted by $[t_1/x_1, \dots , t_n/x_n]$. Moreover, for a sequence $s$ and a label $r$, $s. \sigma$ and $r. \sigma$ are respectively denoted $s[t_1/x_1, \dots , t_n/x_n]$ and $r[t_1/x_1, \dots , t_n/x_n]$.
Every injective substitution which takes values in $V$ is called \emph{renaming}.
Two labels $r_1$ and $r_2$ (respectively two strings $s_1$ and $s_2$) are equal modulo a renaming of variables if there exists a renaming $\sigma$ such that $r_1.\sigma = r_2$ (\resp $s_1.\sigma = s_2$).

Finally, we need another operation on string triplets which allows to combine them together: the string concatenation of $T^\ast$ is noted $\bullet$.
Let $Concat$ be the operation of concatenation on labels which concatenates the three components in the linear order: for  $r \in \Sigma$, $Concat (r) = r_{s} \bullet r_{h} \bullet r_{c}$.

We then have defined a word order structure which encodes specifier/comple\-ment/head relations and two operations (substitution and concatenation). These two operations will be counterparts in the phonological calculus of \textit{merge} and \textit{move}.

\subsubsection{Labelled proofs.}
Before exhibiting the rules of MCG, we need to define the concept of labelling on a subset of rules of the \emph{Minimalist Logic} ($\lto_e$, $\lfrom_e$, $\otimes_e$ and $\sqsubset$).

For a given MCG $G$, let a \emph{$G$-background} be $x: A$ with $x\in V$ and $A \in F$, or $\langle G_1 ; G_2\rangle$ or else $(G_1 , G_2)$ with $G_1$ and $G_2$ some \emph{$G$-backgrounds} which are defined on two disjoint sets of variables.
$G$-backgrounds are series-parallel orders on subsets of $V \times F$.
They are naturally extended to the entropy rule, noted $\sqsubset$.
A \emph{$G$-sequent} is a sequent of the form:
$\Gamma \seq_G (r_s, r_t, r_c) : B$ 
where $\Gamma$ is a $G$-background,
$B \in F$ and $( r_s, r_t, r_c ) \in \Sigma$.

A \emph{$G$-labelling} is a derivation of a $G$-sequent obtained with the following rules:

\begin{center}
\prooftree
	\langle s, A \rangle \in Lex
	\justifies
	\seq_G (\epsilon, s, \epsilon) : A
	\using [Lex]
\endprooftree

\bigskip

\prooftree
	x \in V
	\justifies
	x: A \seq_G (\epsilon, x, \epsilon) : A
	\using [axiom]
\endprooftree

\bigskip

\prooftree
	\Gamma \seq_G r_1 : A \lfrom B
	\quad \Delta \seq_G r_2 : B
	\quad Var(r_1) \cap Var(r_2) = \emptyset
	\justifies
	\langle \Gamma ; \Delta\rangle \seq_G (r_{1s},r_{1t},r_{1c} \bullet Concat(r_2)) : A
	\using [\lfrom_e]
\endprooftree

\bigskip

\prooftree
	\Delta \seq_G r_2 : B
	\quad \Gamma \seq_G r_1 : B \lto A
	\quad Var(r_1) \cap Var(r_2) = \emptyset
	\justifies
	\langle \Gamma ; \Delta\rangle \seq_G  (Concat(r_2)\bullet r_{1s},r_{1t},r_{1c}) : A
	\using [\lto_e]
\endprooftree

\bigskip

\prooftree
	\Gamma \seq_G r_1 : A \otimes B
	\quad \Delta[x: A , y:B] \seq_G r_2 : C
	\quad Var(r_1) \cap Var(r_2) = \emptyset
	\quad A \in \mbox{\textsc{p}}_2
	\justifies
	\Delta[ \Gamma] \seq_G  r_2[Concat(r_1)/x , \epsilon/y ] : C
	\using [\otimes_e]
\endprooftree

\bigskip

\prooftree
	\Gamma \seq_G r : A
	\quad \Gamma' \sqsubset \Gamma
	\justifies
	\Gamma' \seq_G r : A
	\using [\sqsubset] 
\endprooftree
\end{center}

Note that a $G$-labelling is a proof tree of the Minimalist Logic on which sequent hypotheses are decorated with variables and sequent conclusions  are decorated with labels. Product elimination is applied with a substitution on labels and implication connectors with concatenation (a triplet is introduced in another one by concatenating its three components).

\subsection{Lexicon}

MCG encodes informations in the lexicon with types. They are defined over two sets, one of linguistic categories and the other of move features.
Lexical items associate a label and a formula of PCL with respect to the following grammar:

$$
\begin{array}{llll}
\textsc{l} &::= &(\textsc{b}) \lfrom \textsc{p}_1&|\; \textsc{c}\\
\textsc{b}& ::=& \textsc{p}_1 \lto (\textsc{b})&| \;\;\textsc{p}_2 \lto (\textsc{b})\; |\;\; \textsc{c} \;\;|\;\; \textsc{d}\\
\textsc{c} &::= &\textsc{p}_2 \otimes (\textsc{c})&|\;\;\textsc{c}_1\\
\textsc{d} &::= &\textsc{p}_2 \odot (\textsc{d})&|\;\;\textsc{c}_1\\
\textsc{c}_1 &::= &\textsc{p}_1\\
\end{array}
$$

where $\textsc{l}$ is the starting non-terminal and $\textsc{p}_1$ and $\textsc{p}_2$ are atomic formulae belonging to set of features of the MCG (features which trigger \textit{merge} or \textit{move} rules).

Formulae of lexical items get started with a $\lfrom$ as the first connective, and continue with a sequence of $\lto$. 
This corresponds to the sequence of selectors and licensors in MG lexical items. 
These are trigger rule features of MCGs. They give the concrete part of the derivation. 
A formula is ended with an atomic type, the category of the phrase, or a sequence of $\odot$ (which contains at least a specific type, which is also the main category).

For example, the following formula could be the one of an MCG entry:
$(d\lto h \lto j \lto k \lto (a \otimes b \otimes c ))\lfrom m$, whereas this is not :
$(d\lto h \lto j \lto k \lto (a \otimes b \otimes c ))\lfrom m \lfrom p$,
because it has two $\lfrom$.
These formulae have the following structure :
$$(c_m \backslash \ldots \backslash c_1 \backslash (b_1 \otimes \ldots \otimes b_n \otimes a ) ) / d$$ with $a \in$ \textsc{p}$_1$, $b_i \in$ \textsc{p}$_2$, $c_j \in$  \textsc{p} and $d \in$  \textsc{p}$_1$.

The morphism from MG lexicon to MCG ones is defined in \cite{AM07th}. 

\subsection{Rules of MCG}

In the same way as for MG, \cite{Sta97}, rules of MCG are defined over two principles:
\begin{itemize}
\item combining two pieces of derivation: \textit{merge}
\item redefining internal relations in a derivation: \textit{move}
\end{itemize}

As we have mentioned before, MCG is defined over a restriction of PCL: elimination of $\lfrom$ and $\lto$, and $\otimes$.
In the following, in order to distinguish relations in the sequence of hypotheses, a commutative relation will be marked with ',' and a non-one with ';'.

\begin{itemize}
\item \textit{Merge} is the function which combines two pieces of proofs together and it needs an non-com\-mu\-ta\-tive relation to correctly encode relations among words. But in the same application, \textit{merge} will also combine hypothesis of different proofs. And from a linguistic point of view, relations between these hypothesis should be commutative because there is no reason to block access to them.
Then, \textit{merge} combines an elimination of $\lto$ or $\lfrom$ with the application of an entropy rule. 
For the same linguistic reasons as in MG, MCG use two different kinds of \textit{merge} depending on the lexical/non-lexical status of the trigger.

For the word order, \textit{merge} is simply the concatenation of the string of one phrase in the label of the other one (depending of right/left relation):

Lexical trigger:
\begin{center}
\prooftree
\prooftree
	\seq ( r_{s}, r_{h}, r_{c}) : A\lfrom B
	\quad\quad \Delta \seq s : B
	\justifies
	\Delta \seq ( r_{s}, r_{h}, r_{c}\bullet Concat(s)) : A
	\using
	  [\lfrom_e]
\endprooftree
	\justifies
	\Delta \seq ( r_{s}, r_{h}, r_{c}\bullet Concat(s)) : A
	\using
	  [entropy]
\endprooftree

\bigskip

$\Longrightarrow $

\bigskip

\prooftree
	\seq ( r_{s}, r_{h}, r_{c}) : A\lfrom B
	\quad \Delta \seq s : B
	\justifies
	\Delta \seq ( r_{s}, r_{h}, r_{c}\bullet Concat(s)) : A
	\using
	  [mg]
\endprooftree

\end{center}

A \textit{merge} with a lexical item do not explicitly show the order between hypotheses.
But here, the entropy rule is the replacement of a $;$ by a $,$

Non-lexical trigger:
\begin{center}
\prooftree
\prooftree
	 \Delta \seq s : B
	\quad\quad \Gamma \seq (r_{s}, r_{h}, r_{c}) : B \lto A
	\justifies
	\Delta ; \Gamma \seq (Concat(s)\bullet r_{s}, r_{h}, r_{c}) : A
	\using
	  [\lto_e]
\endprooftree
	\justifies
	\Delta , \Gamma \seq (Concat(s)\bullet r_{s}, r_{h}, r_{c}) : A
	\using
	  [entropy]
\endprooftree

\bigskip

$\Longrightarrow$

\bigskip

\prooftree
	 \Delta \seq s : B
	\quad \Gamma \seq (r_{s}, r_{h}, r_{c}) : B \lto A
	\justifies
	\Delta , \Gamma \seq (Concat(s)\bullet r_{s}, r_{h}, r_{c}) : A
	\using
	  [mg]
\endprooftree

\end{center}

\item The encoding of \textit{Move} in MCG is structurally different from the one in MG.
Here, it assumes that a phrase is included in the proof if and only if all its hypotheses are. 
Then, we do not really reinterpret the local tree as in MG, but we directly produce the final derivation tree.
In this way, a \textit{move} is the discharge of hypotheses by a $\otimes$.

For word order, the concatenation of the moved phrase is substituted in the position of the newest hypothesis:

\begin{center}
\prooftree
	\quad \Gamma \seq r_1 : A \otimes B
	\quad  \Delta[u : A, v: B] \seq r_2 : C
	\justifies
	 \Delta[\Gamma] \seq r_2[Concat(r_1) / u , \epsilon /  v] : C
	\using
	  [mv]
\endprooftree

\end{center}

\end{itemize}

Finally, to give account of \cite{St01}, the framework is enriched with rules which modifies the position of the string in the label. There are two kinds of rules:
\begin{itemize}
\item \textit{head movement} where the head of the merged element is concatenated in the final head.
This implies four rules: two to distinguish left and right concatenation and two over the lexical status of the trigger of \textit{merge}. 
\item \textit{Affix hopping} where the head of the trigger is concatenated with the head of the merged element. In the same way, there are four rules to distinguish left from right concatenations and lexical status of the trigger.
\end{itemize}

We use a simple way to encode the different possibilities of merging with $<$ and $>$. Pointing to the connective indicates head-movement and outside defines affix hopping. The lexical status does not need to be represented.
In the following of this paper, only head movement will be used, thus we give this four rules:

\emph{Head-Movement}:
\begin{center}
\prooftree
	 \Gamma \seq (r_{spec}, r_{tete}, r_{comp}) : A\lfrom_{<}B
	\quad \quad \Delta \seq s : B
	\justifies
	\Gamma , \Delta \seq (r_{spec}, r_{tete} \bullet s_{tete}, r_{comp}\bullet Concat(s_{-tete})) :  A
	\using
	  [mg]
\endprooftree

\vspace{2ex}

\prooftree
	 \Gamma \seq (r_{spec}, r_{tete}, r_{comp}) : A_{>}\lfrom B
	\quad \quad  \Delta \seq s : B
	\justifies
	 \Gamma, \Delta \seq (r_{spec}, s_{tete} \bullet r_{tete}  , r_{comp}\bullet Concat(s_{-tete})) : A
	\using
	  [mg]
\endprooftree

\vspace{2ex}

\prooftree
	\Delta \seq  s : B
	\quad \quad \Gamma \seq (r_{spec}, r_{tete}, r_{comp}) : B _{>}\lto A
	\justifies
	\Delta,\Gamma \seq (Concat(s_{-tete})\bullet r_{spec}, s_{tete} \bullet r_{tete}, r_{comp}): A
	\using
	  [mg]
\endprooftree

\vspace{2ex}

\prooftree
	\Delta \seq s : B
	\quad \quad \Gamma \seq (r_{spec}, r_{tete}, r_{comp}) : B \lto_{<} A
	\justifies
	\Delta , \Gamma \seq (Concat(s_{-tete})\bullet r_{spec},  r_{tete} \bullet s_{tete} , r_{comp}) : A
	\using
	  [mg]
\endprooftree
\end{center}

\section{Phases}

\subsection{Encoding Phases in MCG}
\begin{figure}
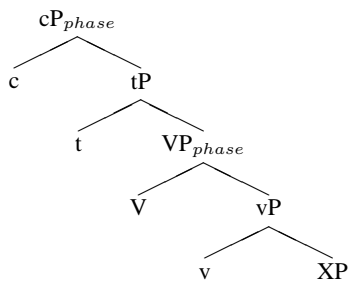

\begin{center}
\leaf{c}
\leaf{t}
\leaf{V}
\leaf{v}
\leaf{XP}
\faketreewidth{\hspace*{30mm}}
\branch{2}{vP}
\faketreewidth{\hspace*{30mm}}
\branch{2}{VP$_{phase}$}
\faketreewidth{\hspace*{30mm}}
\branch{2}{tP}
\faketreewidth{\hspace*{30mm}}
\branch{2}{cP$_{phase}$}
\tree
\end{center}
\caption{Phases in verb structure\label{structurephases}}
\end{figure}

Following \cite{NC99}, Chomsky assumes that the analysis of a sentence is driven by the verb which goes by two specific states:
the \textit{phases} \textit{VP} and \textit{cP}.
Note that neither \textit{tP} nor the decomposition over simple verb form as it is used in usual MCG are \textit{phases} (this is illustrated in figure \ref{structurephases}). Moreover, Chomsky claims that syntactic islands are defined by \textit{phases}. 
That is, the content of a \textit{phase} must be moved to its left-hand side in order to let it accessible. 
This step of the \textit{phase} is called a \textit{transfer}.

According to this structure, items on the right side of the \textit{phase} can not be moved again. This syntactic island is called the Phase Impenetrability Condition (PIC).
The definition of \textit{phases} and the PIC are still under debate.
Nevertheless, an interesting point for MCG is the simultaneous use of commutative and non-commutative properties of the framework to recognize them.

According to Chomsky, a \textit{phase} is a node of the syntactic tree which triggers allowed moves.
Because it is a node, this implies for MCG to combine two subproofs.
It is not possible to simulate this with a single lexical item (which implies a terminal node). Also transfering is the realization of possible \textit{moves} and the direct linking of hypotheses in a \textit{cyclic move}.

Another argument in support of \textit{phases} is in the semantic counterpart. 
This article does not present this second synchronized part of the calculus in MCG. 
Many different works claim that at particular step of the verb derivation some specific thematic roles could be assigned as in \cite{HK93}, \cite{Kra94} and \cite{Bak97}. 
\cite{AM07th} and \cite{ALfg08} describe both arguments for this and the semantic tiers according to these assumptions. But, they do not include the idea of \textit{phases}. 
Only remarks that simplification of derivation requires specific points in which continuation reductions must occur. 
We leave the presentation of consequences of \textit{phases} on semantics to future works.

The analysis of a simple sentence derivation in standard MSG uses 4 items. 
In the following, the verb \textit{read} will be used to illustrate this presentation for a simple active affirmative sentence.

\begin{enumerate}
\item The deep syntactic build of the verb: the verb and all its arguments (except the subject).\label{stepone}
\hfill $\seq v \lfrom_< d$
\item Mode: introduces the subject category and, at least, the \textit{accusative} case.\label{steptwo}

\hfill $\seq k \lto d \lto V \lfrom_< v$
\item Inflection: brings the inflection to the verb. \label{stethree}
\hfill $\seq k \lto t\lfrom_< V$
\item Comp: fully completes the analysis (a question mark, insert in relative clause, etc.)\label{stepfour}
\hfill $\seq c \lfrom t$
\end{enumerate}

In order to keep control on the string associate to the verb, the derivation systematically uses head movement except \textit{comp} which ends the derivation.
We claim that \textit{phases} can be encoded in MCG with a non-commutative order.
The lexical realization of a \textit{phase} item explicitly contains hypotheses in a non-commutative order.

\begin{equation}
\Delta_1, H_1;H_2 \Delta_2 \seq A \label{strucone}
\end{equation}

This order makes a strong boundary in the derivation of the \textit{phase}.
It blocks \textit{move} of elements in complement position to specifier.
These are the only items that are lexically built with hypothesis different from the original definition of MCG \cite{LR01acl}, \cite{AM07th} and \cite{ALR10}.
Using the $\odot_e$ rule of the PCL, the \textit{phase} rule is defined as:

\begin{center}
\prooftree
	 \Delta_{s}, \Delta_{h}, \Delta_{c} \seq (s_{s}, s_{h}, s_{c}) : X \odot Y
	\quad \Gamma_{s}, X;Y, \Gamma_{c} \seq (r_{s}, r_{h}, r_{c}) : Z
	\justifies
	 \Gamma_{s} , \Delta_{s}, \Delta_{h} \seq ( r_{s} \bullet s_{s}, r_{h}, s_h \bullet s_c \bullet r_{c}) : Z
	\using
	  [phase]
\endprooftree
\end{center}

The \textit{phase} rule is the combination of a discharge of hypothesis in non-commu\-ta\-tive order and  a \textit{transfer} step. This \textit{transfer} is the realization of all possible \textit{moves} and parts of \textit{cyclic ones}.
This new rule assembles several individual rules of PCL proof. It may be difficult to follow the derivation step by step. In the following, \textit{phases} are given with more details. Thus, we note $[phase_1]$ the substitution part of the \textit{phase} (the use of $\odot_e$ which combines two proofs) and $[phase_{trans}]$ moves which can be achieved after $[phase_1]$. The main condition to validate a \textit{phase} is $\Delta_c$ and $\Gamma_c$ are empty after $[phase_{trans}]$. This correspond to a phrase in complement position of a \textit{phase} does not stand accessible.
We note MCG$_{phase}$, MCG wit \textit{phases}.

There is a direct consequence of this encoding of \textit{phase} on the structure of the lexical item
Hypotheses on its left-hand side of take the place of the complement of the head. 
Thus, all elements with which it should be combined must be in a specifier position. 
A more complex formula as in (\ref{strucone}) will be built only with $\lto$ :

\begin{equation}
\Delta_1, H_1;H_2 \Delta_2 \seq (s_s, s_h, s_c) B_n \lto \dots \lto B_0 \lto A \label{structwo}
\end{equation}

To take into account of the definition of the \textit{phases} theory and the proposed encoding in MCG, a simple sentence will be divided into two \textit{phases}: one with the \textit{mode} and another with \textit{comp}. Their lexical items are modified in this way into:

\begin{itemize}
\item mode: $\seq k \lto d \lto V \lfrom v$ $\Rightarrow$ $ V;v \seq k \lto d\lto V$
\item comp:  $\seq c \lfrom t$ $\Rightarrow$ $ c;t \seq c$
\end{itemize}

Hypotheses on the left-hand side of formulae receive a straightforward interpretation.
They correspond to the conversion of a proof of a $v$ to a proof of a $V$ (or from a $t$ to a $c$ in the second one). Thus, we need to update the two other formulae in order to combine them with the two previous ones:

\begin{itemize}
\item verb: $\seq v \lfrom d$ $\Rightarrow$  $\seq (V\odot v)\lfrom d$
\item inflection: $\seq k \lto t\lfrom V$ $\Rightarrow$ $\seq k \lto (c \odot t) \lfrom V$
\end{itemize}

Note that the use of $\odot$ in a formula do not imply that this item is one of a \textit{phase}. It means that they take an active part in the construction of the verb. Here, we have extended the main category of the item to the previous one, combined with the category of the following \textit{phase} category with a $\odot$.

In fact, each head item drives a specific part of all elements are included under their relation to the head. But, the unloading of hypotheses (the realization of the \textit{phase}) occurs later. Here, the logical account of the framework updates the derivation in a way that there is no direct intuition with the syntactic tree.
We present  a simple example of interpretation of a \textit{phase} in a proof into a tree: on one hand the starting part of the derivation and on the other the proof which results in the \textit{phase}:

\prooftree
	\seq (A \odot B) \lfrom C
	\quad C \seq C
	\justifies
	C \seq (A \odot B)
	\using
	  [mg]
\endprooftree
\hfill
\prooftree
	A;B \seq D \lto E
	\quad D \seq D
	\justifies
	D, A ; B \seq E
	\using
	  [mg]
\endprooftree

Which correspond to the two following syntactic trees:

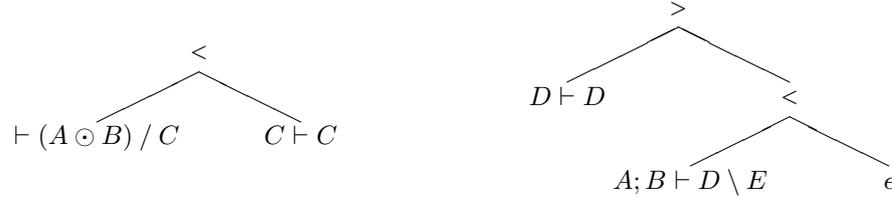
\leaf{$\seq ( A \odot B) \lfrom C$}
\leaf{$C \seq C$}
\faketreewidth{\hspace*{30mm}}
\branch{2}{<}
\tree
\hfill
\leaf{$D \seq D$}
\leaf{$A;B \seq D \lto E$}
\leaf{$\epsilon$}
\faketreewidth{\hspace*{30mm}}
\branch{2}{<}
\branch{2}{>}
\tree

The syntactic tree of the \textit{phase} is not a simple leaf but a tree with an empty position in which the other derivation is substitued:

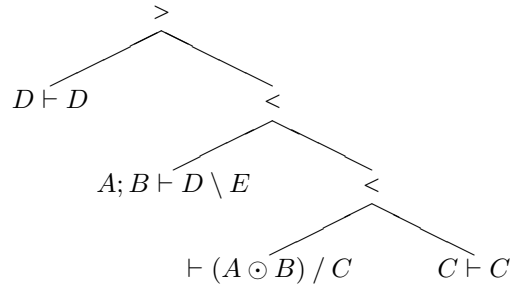
\leaf{$D \seq D$}
\leaf{$A;B \seq D \lto E$}
\leaf{$\seq (A \odot B) \lfrom C$}
\leaf{$C \seq C$}
\faketreewidth{\hspace*{30mm}}
\branch{2}{<}
\faketreewidth{\hspace*{30mm}}
\branch{2}{<}
\branch{2}{>}
\begin{center}
\tree
\end{center}

And it is clear that all information derived in the second tree before the \textit{phase} combination must be updated after.
Finally, for the same reasons as for \textit{merge}, Head Movement and Affix Hopping are needed through the \textit{phase}. The same connectives are used with the same effects. One instance is:

\begin{center}
\prooftree
	 \Delta_{s}, \Delta_{h}, \Delta_{c} \seq (s_{s}, s_{h}, s_{c}) : X \odot_< Y
	\quad \Gamma_{s}, X;Y, \Gamma_{c} \seq (r_{s}, r_{h}, r_{c}) : Z
	\justifies
	 \Gamma_{s} \Delta_{s}, \Delta_{h} \seq ( r_{s} \bullet s_{s}, s_h \bullet r_{h},  s_c \bullet r_{c}) : Z
	\using
	  [phase]
\endprooftree
\end{center}

where the head of the triplet is the concatenation of the two heads.
The following section presents an application of this rules in a full analysis of a simple sentence.

\subsection{Derivation of a simple sentence}
To illustrate derivation with \textit{phases}, this section presents the complete derivation of a simple sentence.
However, extending the analysis to more complex syntactic structures has resulted in defining the lexical entries corresponding with the same principles.
We present the analysis of the following example:
\begin{enumerate}
\item[(1)]The children read a book. \label{simplesentence}
\end{enumerate}

This simple sentence uses the affirmative form, the verb will take the four previous steps. 
In order to build noun phrase with MCG with generative theory, we combine a determiner $\seq (k \otimes d) \lfrom n$ with a noun $\seq n$. It produces a constituent of category $d$ (for \textit{determinal phrase}) which lacks the assignment of syntactic case ($k$).
Then, the following lexicon is used:

$$
\begin{array}{|l|rl|}
\hline
articles &\seq (\epsilon, the, \epsilon) :& k \otimes d \lfrom n\\
 &\seq (\epsilon, a, \epsilon) :&  k \otimes d \lfrom n\\
 \hline
noun &\seq (\epsilon, chlidren , \epsilon) :&    n\\
&\seq (\epsilon, book , \epsilon) :& n\\
 \hline
 
verb & \seq (\epsilon, read , \epsilon) : & (V \odot v)_< \lfrom d\\
(mode) & V; v \seq (\epsilon, \epsilon , \epsilon) : & k \lto d \lto V\\
(inflection) & \seq (\epsilon, - , \epsilon) :  &  k \lto (c \odot t) \lfrom_< V\\
(comp)& c;t \seq (\epsilon , \epsilon , \epsilon) :  &  c \\
\hline
\end{array}
$$

The derivation is a proof, so each main element (the head of each phrases) drives its subproof.
The consequence is that the proof is built in several parts adjusted against each other by the discharge of hypotheses. This property makes the presentation more complex, but the search for proofs (\textit{i.e.} the parsing) could be done in parallel.
In the following, each verb step is presented. Note that the normalization of PCL from \cite{MR07} ensures that we could combine each step one by one as presented here.

\subsubsection{Step 1}
In the first verb step of the derivation, the first position of the verb is saturated by a hypothesis of category $d$. It corresponds to the position of the main component that
occupies the object position in the sentence.
The object is not directly inserted in the derivation because all its features are not yet marked by  hypotheses: the position of $k$ (mark of the syntactic case assignment\footnote{in the semantic part of the calculus, this position is also used for the thematic role assignement.}) is missed. This is a departure from MG, in which all phrases are directly inserted in the derivation.
Here, we only mark positions for insertion.

\begin{center}
\prooftree
	\seq (\epsilon, read , \epsilon) : (V \odot v)_< \lfrom d
	\quad d \seq (\epsilon, u, \epsilon ) : d
	\justifies
	d \seq (\epsilon, read, u) : (V \odot v)
	\using
	  [mg]
\endprooftree
\end{center}

This is the end of the first verb step. The result must be inserted in another proof which contains  hypothesis $V$ and $v$. On one hand, the interpretation of the non-commu\-ta\-ti\-ve relation between $V$ and $v$ in the type could be the saturation of all arguments of an element of type $v$ to produce a $V$. 

\subsubsection{Step 2}

These hypotheses are in the proof driven by the second entry of the verb: \textit{mode} which ends the first \textit{phase}.
First, we need to saturate all positions of \textit{mode} with lexical hypothesis, and then we could:
\begin{enumerate}
\item combine the result with the first verb step,
\item introduce the object of the sentence with a \textit{move} in the transfer part of the \textit{phase}.
\end{enumerate}

The lexical item of \textit{mode} is merged with a hypothesis $k$ and next a $d$:
\begin{center}
\prooftree	
	\quad d \seq (\epsilon, w, \epsilon ) : d
	\prooftree
		k \seq (\epsilon, v, \epsilon ) : k
		\quad  V; v \seq (\epsilon, \epsilon , \epsilon) : k \lto d \lto V
		\justifies
		k , V; v \seq (v , \epsilon, \epsilon ) : d \lto V
		\using
		  [mg]
	\endprooftree
	\justifies
	d, k ,V; v \seq (w\; v, \epsilon, \epsilon ) : V
	\using
		  [mg]
\endprooftree
\end{center}

At this point of the derivation, this result is combined with the first verb step with a $[phase]$ with head movement. Note that the string of the head of the discharged is concatenated to the string of the head position in order to keep all structural information over the verb accessible to the full derivation.
The substitution part of the \textit{phase} produces:
\begin{center}
\prooftree
	d \seq (\epsilon, read, u) : (V \odot v)_<
	\quad d, k ,V; v \seq (w\; v, \epsilon, \epsilon ) : V
	\justifies
	d, k , d \seq (w\; v,read, u ) : V
	\using
	  [phase_1]
\endprooftree
\end{center}

Before ending the \textit{phase}, we perform a \textit{move} with all positions of the utterance's object (they are now in the proof). Then it could be introduced in the derivation with a transfer. 
In parallel, the determiner phrase is built with the two lexical entries "a" and "book" by a \textit{merge}:

\begin{center}
\prooftree
	\seq (\epsilon, a, \epsilon):  (k \otimes d) \lfrom n
	\quad \seq (\epsilon, book , \epsilon):  n
	\justifies
	\seq (\epsilon , a , book): k \otimes d 
	\using
	  [mg]
	\thickness=0.07em
\endprooftree
\end{center}

Thereby, it is discharged in the main proof. The choice of the $d$ with which to carry out the unloading is not left by chance. The derivation must be continued with the one which empties the previous verb step with a \textit{move}. This is exactly the interpretation of transfer part in \cite{NC99}.

\begin{center}
\prooftree
	\seq (\epsilon , a , book): k \otimes d 
	\quad d, k , d \seq (w\; v,read, u ) :  V
	\justifies
	d \seq (w\; a \; book, read, \epsilon ) : V
	\using
	  [phase_{trans}]
	\thickness=0.07em
\endprooftree
\end{center}

This \textit{move} substitute in the newest variable as define in \cite{AM07th}.
It is the full realization of the constituent.
In this \textit{phase}, non-commutativity and commutativity are both used.
Non-commutativity in order to keep the structure of the verb and commutativity to unload hypotheses of the nominal phrase.
This underlies the assumption that the order of the features of the noun could not be presupposed.
This is reinforced by the analysis of questions where that object constituent undergoes one more \textit{move} and then must be explicitly transfered from right to left part of the \textit{phase}.
Now, it is the end of the second verb step. The derivation continues by preparing the third one.

\subsubsection{Step 3}

This part of the derivation must be combined with the next lexical entry of the verb, the \textit{inflection}.
In this part of the verb step, it was merged with the previous result and next with a $k$ hypothesis - the position of the subject case:

\begin{center}
\prooftree
	k \seq (\epsilon, z, \epsilon) : k
	\prooftree
		\seq (\epsilon, - , \epsilon) :  k \lto (c \odot t) \lfrom_< V
		\quad d \seq (w\; a \; book, read,  \epsilon) : V
		\justifies
		d \seq (\epsilon, read,w\; a \; book) : k \lto (c \odot t)
		\using
		  [mg]
		\thickness=0.07em
	\endprooftree
	\justifies
	k, d \seq (z, read,w \; a \; book) : (c \odot t)
	\using
	  [mg]
	\thickness=0.07em
\endprooftree
\end{center}

This allows to discharge hypothesis about the subject constituent which it also build:

\begin{center}
\prooftree
	\seq (\epsilon, the, \epsilon):  (k \otimes d) \lfrom n
	\quad \seq (\epsilon, children , \epsilon):  n
	\justifies
	\seq (\epsilon , the, children): k \otimes d 
	\using
	  [mg]
	\thickness=0.07em
\endprooftree
\end{center}

And unloaded:

\begin{center}
\prooftree
	\seq (\epsilon , the, children): k \otimes d 
	\quad k, d \seq (z, read, w \; a \; book) : (c \odot t)
	\justifies
	\seq (the \; children, read, a \;book) : (c \odot t)
	\using
	  [mv]
	\thickness=0.07em
\endprooftree
\end{center}

\subsubsection{Step 4}

This example stands for a very simple sentence, then the last verb step corresponds only to the combination of the  current bypass with the lexical entry \textit{comp} by a \textit{phase} with nothing in the transfer part:

\begin{center}
\prooftree
	\seq (the \; children, read, a \;book) : (c \odot t)
	\quad c;t \seq (\epsilon, \epsilon, \epsilon) : c
	\justifies
	\seq (\epsilon, \epsilon, the \; children\; read\; a \;book) : c
	\using
	  [phase]
	\thickness=0.07em
\endprooftree
\end{center}

This ends the last \textit{phase} and thus the derivation. The proof matches the string \emph {The children read a book}.
An important distinction with the previous versions of these grammars is in the use of lexical item without phonological part. Here, only the lexical items used in the \textit{phase} process are necessary, but the structure of items for \textit{phases} imposes a strict order in their pooling.

\subsection{Question}
In the previous example, the transfer part of the two \textit{phases} is not really efficient.
The first one introduced the object of the utterance and the second only ended the derivation.
For questions, the $comp$ item is more complex because it introduces the last feature of the object.
This time, its lexical item is :  $c;t \seq (\epsilon , \epsilon , \epsilon) :  wh \lto c $. And it is only afterward that a hypothesis $wh$ is introduced that the object could be introduced in the derivation.
But it means that in the previous \textit{phase}, the constituent mark must be transferred from the left to the right part of the first \textit{phase}.
This is done with a \textit{cyclic move}: the introduction of a new hypothesis $k\otimes d \seq k\otimes d $, which explicitly connect the two hypotheses.

In the lexicon, only $comp$ is modified and an item for which is added :
$$which \seq (\epsilon, \epsilon, \epsilon) : (wh \otimes (k \otimes d))$$

The derivation before the \textit{phase} is still the same.
\begin{enumerate}
\item First step procedure:
\begin{center}
\prooftree
	\seq (\epsilon, read , \epsilon) : (V \odot v)_< \lfrom d
	\quad d \seq (\epsilon, u, \epsilon ) : d
	\justifies
	d \seq (\epsilon, read, u) : (V \odot v)
	\using
	  [mg]
\endprooftree
\end{center}

\item Saturation of positions of $mode$:
\begin{center}
\prooftree	
	\quad d \seq (\epsilon, w, \epsilon ) : d
	\prooftree
		k \seq (\epsilon, v, \epsilon ) : k
		\quad  V; v \seq (\epsilon, \epsilon , \epsilon) : k \lto d \lto V
		\justifies
		k , V; v \seq (v , \epsilon, \epsilon ) : d \lto V
		\using
		  [mg]
	\endprooftree
	\justifies
	d, k ,V; v \seq (w\; v, \epsilon, \epsilon ) : V
	\using
		  [mg]
\endprooftree
\end{center}
\item Construction of the object constituent:
\begin{center}
\prooftree
	\seq (\epsilon, which, \epsilon):  (wh \otimes (k \otimes d)) \lfrom n
	\quad \seq (\epsilon, book , \epsilon):  n
	\justifies
	\seq (\epsilon , which , book): (wh \otimes (k \otimes d))
	\using
	  [mg]
	\thickness=0.07em
\endprooftree
\end{center}
\end{enumerate}

We get all the necessary material to process the \textit{phase}.
Its first part combines:
\begin{center}
\prooftree
	d \seq (\epsilon, read, u) : (V \odot v)
	\quad d, k ,V; v \seq (w\; v, \epsilon, \epsilon ) : V
	\justifies
	d, k , d \seq (w\; v,read, u ) : V
	\using
	  [phase_1]
\endprooftree
\end{center}

In the treatment of this utterance, the transfer part is not able to directly discharge the two hypotheses of the determiner phrase.
A \textit{cyclic move} is used in order to store the access to this element. At the same time, the \textit{phase} move it on its left part:

\begin{center}
\prooftree
	k \otimes d \seq (\epsilon, W, \epsilon):k \otimes d
	\quad d, k , d \seq (w\; v,read, u ) : V
	\justifies
	d, k \otimes d \seq (w\; W,read, \epsilon ) : V
	\using
	  [phase_{trans}]
\endprooftree
\end{center}

The derivation continues with the same third step and produces:

$$k \otimes d \seq (the \; children, read, W) : (c \odot t)$$

Finally, before the last \textit{phase}, the derivation introduces a $wh$ hypothesis which will allow the \textit{move} of object after the first part of the \textit{phase} realization. The \textit{move} of the transfer part is:

\begin{center}
\prooftree
	\seq (\epsilon , which , book): (wh \otimes (k \otimes d))
	\quad wh, k \otimes d \seq (y \, the \, children, read, W) : c
	\justifies
	\seq (which \, book, \epsilon, \, the \, children \, read) : c
	\using
	  [phase_{trans}]
\endprooftree
\end{center}

In this example, only the transfer in the first \textit{phase}, which accounts for the \textit{cyclic move} of the constituent allows to complete the derivation.

\subsection{Blocked derivation with PIC}

A very important point in the definition of the \textit{phase} rule is the fact that the complementizer part of hypothesis must be removed. This property encodes the Phase Impenetrability Condition. 
The previous one does not contain such problem because the transfer part of the first \textit{phase} achieves all \textit{moves} which empty complementizer hypotheses.

A simple example extracted from the previous one is the case where the $k$ hypothesis in the second step of the verb is not included. Thus the derivation must failed because one hypothesis is away.
The lexical entry corresponding to $mode$ is:

\begin{center}
$V; v \seq (\epsilon, \epsilon , \epsilon) :  d \lto V$
\end{center}

which produces a conclusion of a proof of type $V$ with only $d$ in the left hand side:
\begin{center}$ d, V; v \seq (w \epsilon, \epsilon ) : V$
 \end{center}
The result of the first part of the \textit{phase} is:

\begin{center}
\prooftree
	d \seq (\epsilon, read, u) : (V \odot v)
	\quad d,V; v \seq (w\; v, \epsilon, \epsilon ) : V
	\justifies
	d, d \seq (w\; v,read, u ) : V
	\using
	  [phase_1]
\endprooftree
\end{center}

And the transfer part does not contribute to this step. The part of the $\Gamma_{c}$ of the \textit{phase} rule is not removed.
The structure of the proof which blocks the derivation is the case where the constituent is in complementizer position of the head.
 
We would remark that derivations with \textit{phases} immediately block the process unlike traditional MCG or MG which perform the full derivation before concluding that a specific feature stand at the end (and reject the derivation).
 
Even if this example is quite simple, it shows that the encoding of PIC directly uses properties of the MCG. 
Unlike the other constraints, we do not need to propose new rules. 
That insure to keep the same generative power for MCG$_{phase}$.
The derivation strictly controls the structure and check internal relations.

\section{Conclusion}

The main aim of this paper is to introduce the concept of \textit{phase} from minimalism into type logical grammars, simulating the generative theory of Chomsky which has been an open question since \cite{NC99}.
It involves the introduction of a new rule into the system, and highlights commutative and non-commutative relationships between elements of the parsing process.
Moreover, this addition is not \textit{ad hoc} as it allows full use of the properties of PCL underlying the formalism.
This new rule is the composition of a substitution of hypotheses in commutative relations, followed by a transfer that is either the realization of a \textit{move} became possible, or a \textit{cyclic move}.
This proposal also involves a new linguistic interpretation of \textit{cyclic move}.

A full description of the system would require additional details \cite{AM07th}.
However, we emphasize the role played by \textit{phases} at the syntactic level to define \textit{islands} where the encoding of PIC is simply the use of logical properties of the framework.
Furthermore, we claim that the use of \textit{phases} at semantic level corresponds to the introduction of thematic role predicates (variables related to the reification of formulas and substitution of variables). They also mark points in the semantic tiers where the context must be reduced. It plays a crucial role at the semantic level by marking reduction point for continuations.

The introduction of these \textit{phases} confirms the use of a logical system simultaneously handling relations commutative and non-commutative at plays in linguistic analysis. 
Distributing the properties on each component used in this analysis can produce fine performances
A remaining issue for this description is the formalization of the \textit{phase} as reduction point at the semantic level that would reduce ambiguities of scope of quantifiers.
In addition, the study of equivalence between the MCG with \textit{phases} and MG remains an open question.

\section*{Acknowledgments}

The author would like to express his gratitude to reviewers for their precise remarks, Corinna Anderson, Sai Qian and Sandrine Ribeau for their readings.

%%================================================================
\end{document}